\documentclass[letterpaper]{article} 
\usepackage{aaai24}  
\usepackage{times}  
\usepackage{helvet}  
\usepackage{courier}  
\usepackage[hyphens]{url}  
\usepackage{graphicx} 
\usepackage{natbib}  
\usepackage{caption} 
\usepackage{algorithm}
\usepackage{algorithmicx}
\usepackage{algpseudocode}

\usepackage{newfloat}
\usepackage{listings}
\DeclareCaptionStyle{ruled}{labelfont=normalfont,labelsep=colon,strut=off} 
\floatstyle{ruled}
\newfloat{listing}{tb}{lst}{}
\floatname{listing}{Listing}
\usepackage{multirow}
\usepackage{amsmath}
\usepackage{microtype}
\usepackage{diagbox}
\usepackage{graphicx}
\usepackage{color}
\usepackage{array}
\usepackage{float}
\usepackage{multirow}
\usepackage{mathtools}
\usepackage{amssymb} 
\usepackage{pifont}

\title{Manifold-Based Verbalizer Space Re-embedding for Tuning-Free Prompt-Based Classification}
\author {
    Haochun Wang, Sendong Zhao\thanks{Corresponding author}, Chi Liu, Nuwa Xi, Muzhen Cai, Bing Qin, Ting Liu
}

\affiliations {
   Research Center for Social Computing and Information Retrieval, Harbin Institute of Technology, China\\
    \{hcwang, sdzhao\}@ir.hit.edu.cn
}

\usepackage{bibentry}

\begin{document}

\maketitle

\begin{abstract}
Prompt-based classification adapts tasks to a cloze question format utilizing the $\mathtt{[MASK]}$ token and the filled tokens are then mapped to labels through pre-defined verbalizers. Recent studies have explored the use of verbalizer embeddings to reduce labor in this process. However, all existing studies require a tuning process for either the pre-trained models or additional trainable embeddings. Meanwhile, the distance between high-dimensional verbalizer embeddings should not be measured by Euclidean distance due to the potential for non-linear manifolds in the representation space. In this study, we propose a tuning-free manifold-based space re-embedding method called \textbf{\underline{L}}ocally \textbf{\underline{L}}inear \textbf{\underline{E}}mbedding with \textbf{\underline{I}}ntra-class \textbf{\underline{N}}eighborhood \textbf{\underline{C}}onstraint (\textbf{LLE-INC}) for verbalizer embeddings, which preserves local properties within the same class as guidance for classification. Experimental results indicate that even \textit{without tuning any parameters}, our LLE-INC is on par with automated verbalizers with parameter tuning. And with the parameter updating, our approach further enhances prompt-based tuning by up to 3.2\%. Furthermore, experiments with the LLaMA-7B, 13B and 65B indicate that LLE-INC is an efficient tuning-free classification approach for the hyper-scale language models.

\end{abstract}
\section{Introduction}

Large language models have seen remarkable success in natural language processing (NLP) with pre-training on vast amounts of unlabeled data via masked language model (MLM) \cite{devlin-etal-2019-bert, liu2019roberta} and fine-tuning for the downstream tasks \cite{howard-ruder-2018-universal}. Despite these advancements, the above paradigm necessitates a substantial amount of labeled training data, which poses challenges when attempting to train an additional classification layer during fine-tuning with limited data \cite{liu2021pre}. 

\begin{figure}
     \centering
    \includegraphics[width=0.98\columnwidth]{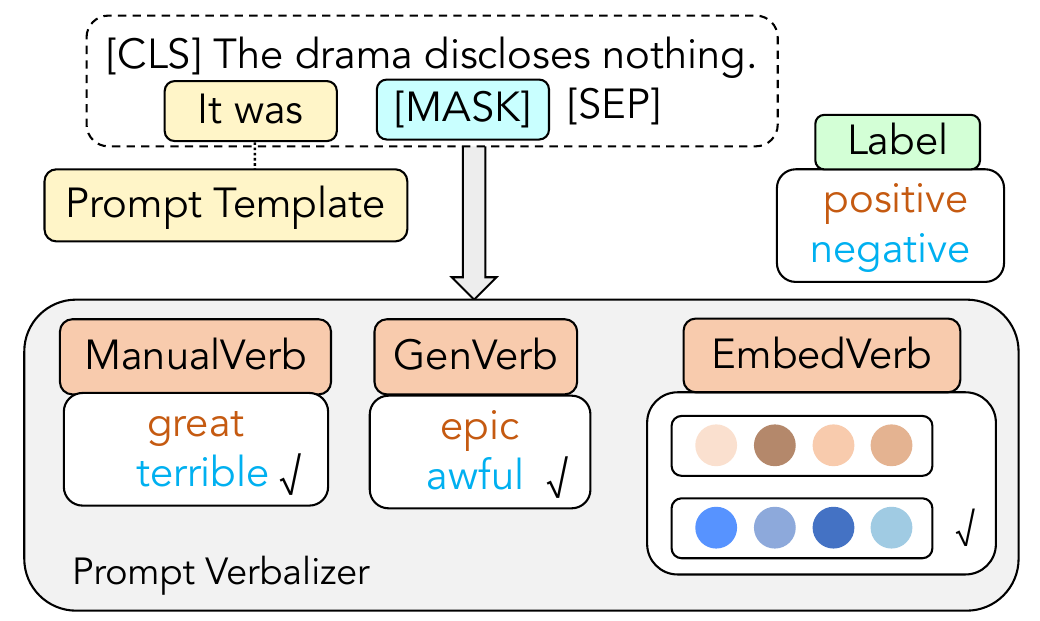}
\caption{Illustration of three typical verbalizer design methods for prompt-based tuning.}
    \label{figure:verbalizer}
\end{figure}

Prompt-based tuning is a method that addresses this issue by converting the task into a cloze question \cite{schick2021exploiting}, where the pre-trained models (PTMs) are prompted to fill in the $\mathtt{[MASK]}$ token with a suitable token from the vocabulary list, similar to the pre-training task. 

Only a limited number of tokens from the vocabulary list are chosen as \textit{verbalizers}, which map each token to its corresponding class, as illustrated in Figure \ref{figure:verbalizer}. Furthermore, variations in the verbalizer designs for a task can significantly impact the performance of prompt-based tuning \cite{gao-etal-2021-making}. 
As shown in Figure \ref{figure:verbalizer}, recent studies on prompt-based tuning have employed verbalizers in three distinct ways: manually selected token verbalizers (ManualVerb) by domain experts empirically \cite{schick2021exploiting}, automatically generated token verbalizers (GenVerb) through gradient-based searching or language model \cite{shin2020eliciting, gao-etal-2021-making}, and embedded verbalizers (EmbedVerb) through optimizing trainable embeddings \cite{zhang2022differentiable, hambardzumyan2021warp, cui2022prototypical}.

\begin{figure}
     \centering
    \includegraphics[width=0.98\columnwidth]{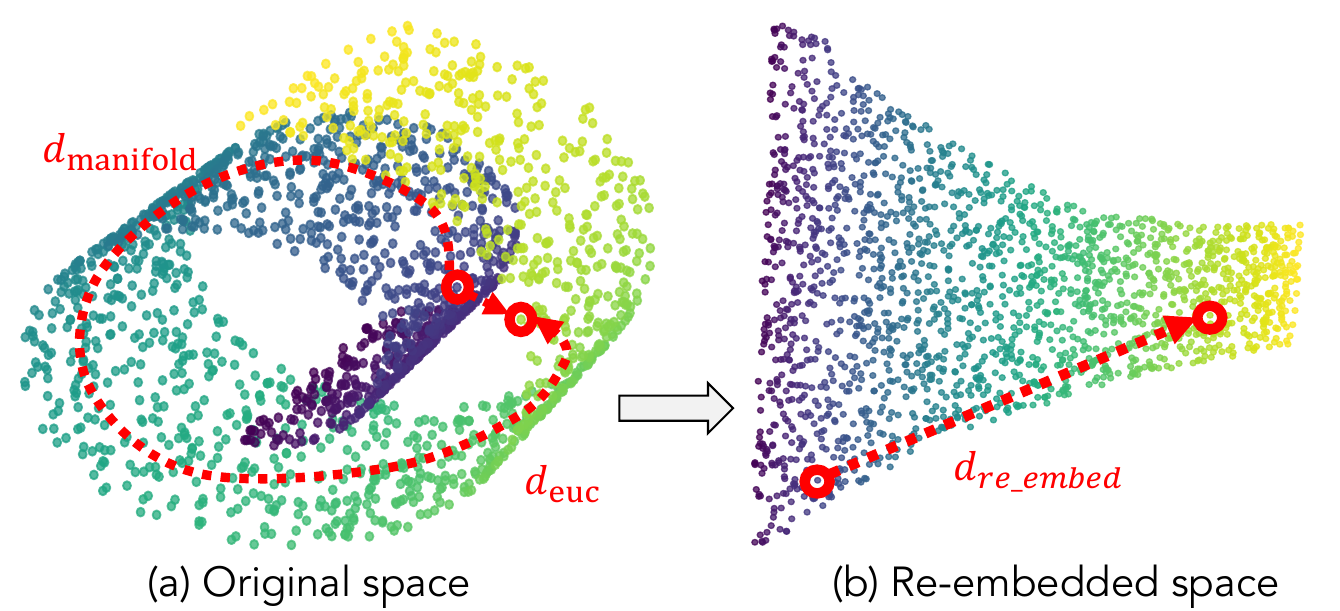}
    \caption{A sketch map for the space re-embedding. $d_{euc}$ is the Euclidean distance in the original space. $d_{manifold}$ is the distance along the manifold shape in the original space. $d_{re\_embed}$ is the Euclidean distance in the re-embedded space.}
    \label{figure:manifold_illustration}
\end{figure}

There are two remaining challenges for the prompt-based classification: (1) All the above three kinds of verbalizers require \textit{updating the parameters} either in the PTMs or the training of extra embeddings, which requires great computational resources, especially for the large language models (LLMs). (2) The \textit{potential manifold} comprising the verbalizer embeddings distributed in the high-dimensional space has not been taken into account and the intra-class neighborhood relationship has been ignored. While a wide range of applications treat the observed space in the PTM as a high-dimensional Euclidean space and utilize Euclidean distance to measure the distance between different vectors \cite{yu2021fine, gao2021simcse}, this may not be as appropriate for the verbalizer embedding on a manifold which is a topological space that only resembles Euclidean space locally \cite{lee2010introduction}. For instance, in Figure \ref{figure:manifold_illustration}, given two 3-D data points marked with red circles, the Euclidean distance between them is $d_{euc}$ but it is clear that when considering the manifold, the distance along the roll $d_{manifold}$ can more accurately depict the relationship between the points, with $d_{euc}\ll d_{manifold}$. Similarly, verbalizer embeddings with even higher dimensions can be also distributed on highly distorted manifolds. Manifold learning can estimate the distance between nearby points with the local neighbors and that between distant points with multi-hop neighbors along the manifold shape, which helps to alleviate this issue. 

In this study, we propose manifold-based space re-embedding for tuning-free prompt-based classification. Motivated by the Locally Linear Embedding \cite{roweis2000nonlinear}, we introduce the Locally Linear Embedding with Intra-class Neighborhood Constraint (LLE-INC) for the calibration of $\mathtt{[MASK]}$ embeddings in the verbalizer space. We posit that the embeddings of the $\mathtt{[MASK]}$ token from the instances that share the same class should be close to one another on a particular manifold. Thus, the LLE-INC aims to reconstruct the representation space into a new Euclidean space, where the linear relationship between data points in the same class is consistent with that in the original space. The recovered Euclidean metric space can improve the distance metrics on high-dimensional verbalizer embeddings, and thus empower the prompt-based PTMs without any tuning on the parameters in the PTMs or the addition of new trainable parameters. The above tuning-free paradigm also explores the potential of LLMs for classification tasks with labeled instances. \footnote{We release our code at https://github.com/SCIR-HI/LLE-INC.}

In summary, our contributions are as follows:
\begin{itemize}
    \item We address a significant challenge in the use of embedded verbalizers for prompt-based classification, specifically the potential for Euclidean distance metrics in the original verbalizer space to ignore the existence of manifolds in high-dimensional space.
    \item We propose the tuning-free LLE-INC method to re-embed the verbalizer space into a new, recovered space, leveraging the intra-class neighborhood relationship in the few-shot datasets. We evaluate the LLE-INC on 10 benchmark datasets. Without tuning any parameters, our approach can produce results that are on par with or superior to baselines. Moreover, when combined with the tuning of PTMs with contrastive learning, our approach outperforms the baselines consistently.
    \item We explore a novel and efficient strategy of leveraging the output embeddings of PTMs with no parameter updating for tuning-free applications for LLMs.
\end{itemize}

\section{Related Work}

\paragraph{Prompt-based Tuning}
PTMs have demonstrated effectiveness in various NLP tasks recently \cite{peters-etal-2018-deep, devlin-etal-2019-bert, liu2019roberta}. However, in few-shot settings, fine-tuning PTMs with the additional task-specific linear layer can be challenging. To address this issue, prompt-based tuning has been proposed to adapt downstream tasks to the pre-training paradigm by converting them into a cloze-question format using a prompt template and prompt verbalizer \cite{schick2021exploiting,gao-etal-2021-making,du-etal-2023-make}, as illustrated in Figure \ref{figure:verbalizer}. The PTMs are expected to predict these verbalizers for the $\mathtt{[MASK]}$ token, which are then mapped to the corresponding labels. Early attempts used a prompt template and verbalizer selected manually through domain knowledge \cite{schick2021exploiting,wang-etal-2022-prompt}. Subsequently, Gao et al. \cite{gao-etal-2021-making} and Liu et al. \cite{liu2021gpt} proposed methods for searching for the optimal prompt. Additionally, some works have focused on continuous prompt engineering, which involves freezing the parameters of PTMs and only updating the prompt embedding. Lester et al. \cite{lester2021power} and Liu et al. \cite{10.1007/978-3-031-44693-1_51} replaced the prompt template with trainable embeddings and optimized the prompt embeddings with extra training steps. Li et al. \cite{li-liang-2021-prefix} applied the continuous prompt to generation tasks. In this study, we focus on the discrete prompt template in natural language which is feasible with the tuning-free setting. 

\paragraph{Prompt Verbalizer Engineering}
Prompt verbalizers map the outputs of the Masked Language Model (MLM) task of PTMs to actual labels \cite{liu2021pre}, and the settings for these verbalizers can significantly impact model performance \cite{gao-etal-2021-making}. Schick et al. \cite{schick2021exploiting, schick2021s} used manually designed, task-specific prompt verbalizers, which require domain expertise and a vast amount of time to obtain the optimal verbalizers. To address this problem, Gao et al. \cite{gao-etal-2021-making} and Shin et al. \cite{shin2020eliciting} proposed generating prompt verbalizers from the vocabulary list by maximizing the conditional probability. However, this approach may result in generated verbalizers that are not coherent in the context \cite{shin2020eliciting}. Hu et al. \cite{hu2022knowledgeable} expanded the candidate verbalizer list with tokens that share semantic similarity via external knowledge bases. Recently, Cui et al. \cite{cui2022prototypical} proposed prototypical verbalizer embeddings, which are learned class prototypes based on training instances. It can achieve state-of-the-art performance for automating verbalizer design, but it is still less effective than elaborate manual verbalizers.

\paragraph{Manifold Learning}
As data in high-dimensional space can be distributed on a specific non-linear manifold, it can be unreasonable to measure differences using Euclidean distance. To address this problem, manifold learning transforms the data in the original space into a new space, based on the assumption that local manifold space is homeomorphic to Euclidean space  Roweis et al. \cite{roweis2000nonlinear} proposed the locally linear embedding to accomplish non-linear dimension reduction via preserving local properties of high dimensional data in the space reconstruction to exploit local symmetries. Inspired by this, Hasan et al. \cite{hasan2017word} re-embedded word embeddings by converting pre-trained vectors to a new Euclidean space with word frequency ranking. Chu et al. \cite{yongherefining} proposed a dynamic word selection method to address the singularity problem in the matrix. Wang et al. \cite{wang2022manifold} integrated manifold-based geometric structures and refined sentence embeddings. In this study, we apply manifold learning to prompt learning and leverage the intra-neighborhood relationship in space re-embedding. 

\section{Method}
\begin{figure*}[t]
    \centering
    \includegraphics[width=0.95\textwidth]{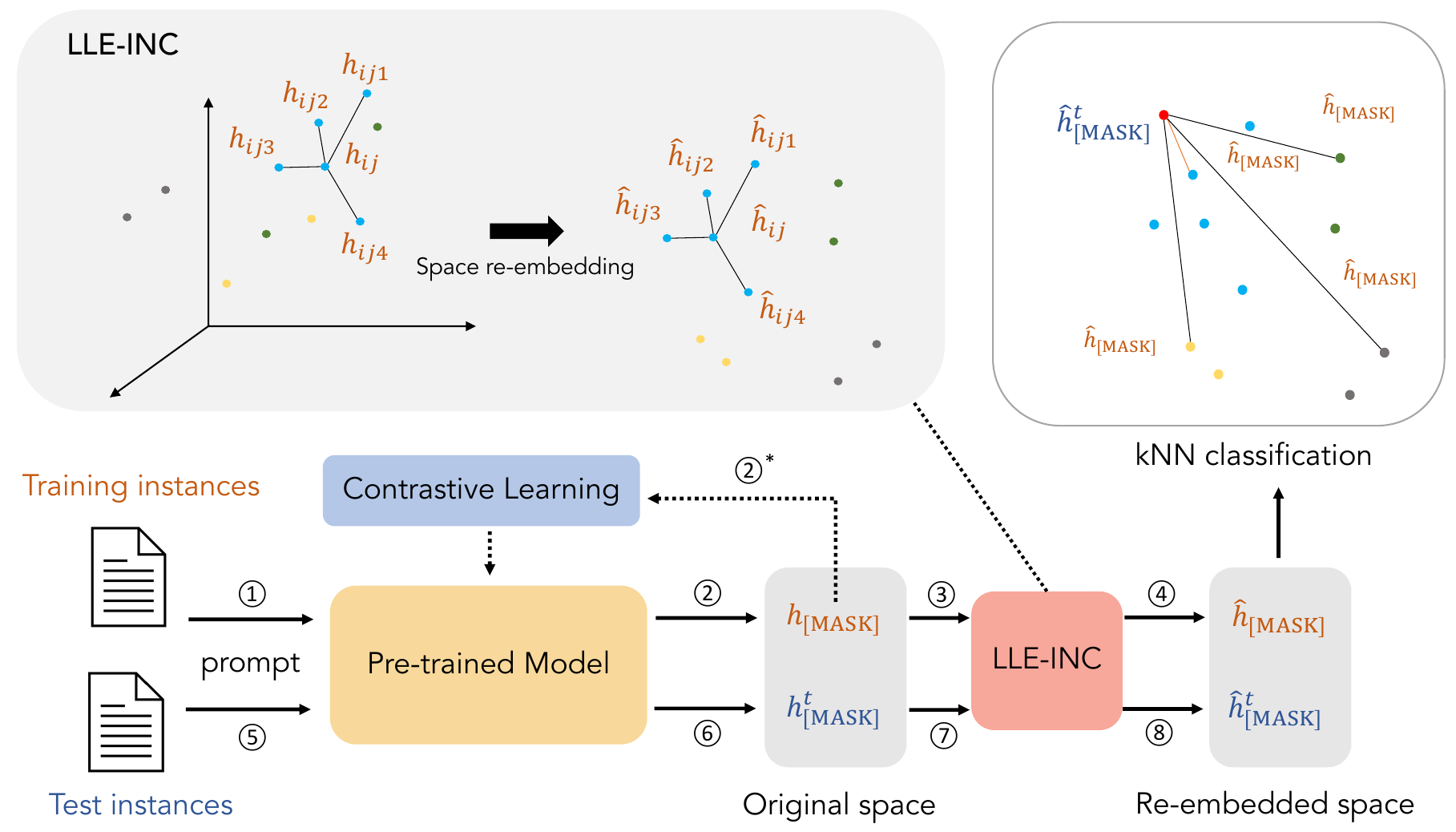}
    \caption{Manifold-based verbalizer space re-embedding for prompt-based tuning. LLE-INC re-embeds the representation space based on the intra-neighbor constraint within the training instances and a kNN classifier makes predictions with the re-embedded representation. The contrastive learning (\ding{173}$^{*}$) is a supplementary module and is not essential.}
    \label{figure:model}
\end{figure*}
\subsection{Continuous Verbalizer Embedding}
In the previous section, we presented an overview of the general process of prompt-based tuning with discrete verbalizers. Manually elaborating verbalizers, particularly for multi-class classification tasks, can be a time-consuming and labor-intensive process. Inspired by \cite{jiang2022promptbert, cui2022prototypical}, we utilize the $\mathtt{[MASK]}$ token embeddings as the representation of instances and classify test instances based on the embedding distance.

\begin{equation}
    h_{\mathtt{[MASK]}} = h_{\text{instance}}
\end{equation}

Since the Euclidean distance between embeddings may not be able to capture the potential manifolds in the embedding space, in the following section, we describe how we adopt the method of manifold-based space re-embedding to re-embed the representation of verbalizers with the prompt-based tuning of PTMs.

\subsection{Manifold-based Re-embedding}
Our manifold-based re-embedding methodology is illustrated in Figure \ref{figure:model}. We begin with the output embeddings of the $\mathtt{[MASK]}$ token in the original space from the PTM and proceed through the following steps to obtain the re-embedded representation. In step 1, the distributed representations of $\mathtt{[MASK]}$ tokens are gathered from all classes in the training set. In step 2, we fit the manifold learning model of \textbf{\underline{L}}ocally \textbf{\underline{L}}inear \textbf{\underline{E}}mbedding with \textbf{\underline{I}}ntra-class \textbf{\underline{N}}eighborhood \textbf{\underline{C}}onstraint (\textbf{LLE-INC}) to the collected $\mathtt{[MASK]}$ embeddings, allowing for the re-embedding of representation from the original space into a new one. In step 3, the fitted model transforms the $\mathtt{[MASK]}$ embeddings in the test instances into the re-embedded space. Finally, in step 4, we apply the k-nearest neighbors (kNN) algorithm to the $\mathtt{[MASK]}$ embeddings in test instances for the classification tasks. For easier understanding, a table outlining the notation for the variables used in this study can be found in Appendix \ref{appendix:notation}.

\paragraph{Step 1.} In the few-shot scenario, we have a $N$-way $k$-shot setting for the training set. The training samples are provided as inputs to the PTM wrapped with prompts. We collect the output hidden state representation of $\mathtt{[MASK]}$ tokens $h_{ij},\ i\in1,2,...,N;j\in1,2,...,k$ as the embeddings in the original PTM space, which is in high dimension.  

\paragraph{Step 2.} 
Since the metric based on Euclidean distance is unaware of the possible manifold in the original space, we leverage the original embeddings in Step 1 to fit a manifold-based model. Locally Linear Embedding (LLE) \cite{roweis2000nonlinear} assumes that the local space is homeomorphic to the Euclidean space and a data point is a linear combination of its neighboring points. Thus, a data sample $x_i$ can be reconstructed by the linear combination of its $K$-nearest neighbor points. 

\begin{equation}
    x_i = w_{ij}x_{j} + w_{ik}x_{k} + w_{il}x_{l} + ...... 
\label{formula:linear-combination}
\end{equation}

It is expected that the linear relationship in Formula \ref{formula:linear-combination} with the neighbors can be preserved following reconstruction. However, due to the limited data scale, $K$-nearest neighbors may not accurately represent the shape of the manifold in the high-dimensional space, as the training instances may be distributed sparsely in the original space. Consequently, we assume that the $\mathtt{[MASK]}$ embeddings of the instances within the same class should be situated proximally on a specific manifold in the original space and that the relationship between an arbitrary embedding and its intra-class neighbors should be maintained through the verbalizer space reconstruction. With this assumption, a Locally Linear Embedding with Intra-class Neighborhood Constraint (LLE-INC) model is proposed in order to preserve the linear relationship between intra-class neighbors. 

The weight coefficient $w_{ijm}$ is determined by locating the $c$-nearest neighbors of each $\mathtt{[MASK]}$ embedding $h_{ij}$, that are $h_{ijm}$, which share the same class with $h_{ij}$ and minimizing the following error function. 
The sum-to-one constraint in Formula \ref{formula:minimization}, which reflects the intrinsic geometric relationship with the corresponding neighbors \cite{saul2000introduction}, ensures that embeddings are transformation-invariant.

\begin{equation}
    \min \sum_{i=1}^{N} \sum_{j=1}^{k} \left\| h_{ij}-\sum_{m=1}^{c} w_{ijm} h_{ijm} \right\| ^{2}\\
\label{formula:minimization}
\end{equation}

\begin{equation*}  
    \text {s.t.} \sum_{m=1}^{c} w_{ijm}=1
\end{equation*}

Since $\sum_{m=1}^{c} w_{ijm}=1$, we have $\textbf{w}_{ij} \textbf{I}=1$, where $\textbf{I}\in \mathcal{R}^{c\times 1}$ is an all-ones column vector. The objective function can be written in the format of vector, where $\textbf{h}_{ij} \in \mathcal{R}^{1\times d}$, $\textbf{w}_{ij} \in \mathcal{R}^{1\times c}$, $\textbf{H}_{ij} \in \mathcal{R}^{c\times d}$.

\begin{equation}
 \begin{aligned}
    f &= \left\| h_{ij}-\sum_{m=1}^{c} w_{ijm} h_{ijm} \right\| ^{2}\\
    &= \left\|\textbf{h}_{ij}-\textbf{w}_{ij}\textbf{H}_{ij} \right\|^{2} \\
    &= \left\|\textbf{w}_{ij} \textbf{I}\textbf{h}_{ij} - \textbf{w}_{ij}\textbf{H}_{ij} \right\|^{2} \\
    &= \left\|\textbf{w}_{ij} (\textbf{I}\textbf{h}_{ij} - \textbf{H}_{ij}) \right\|^{2} \\
    &= \textbf{w}_{ij} (\textbf{I}\textbf{h}_{ij} - \textbf{H}_{ij})(\textbf{I}\textbf{h}_{ij} - \textbf{H}_{ij})^T  \textbf{w}_{ij}^T \\
 \end{aligned}
\end{equation}

With the Lagrangian multiplier, we can obtain the optimal $\textbf{w}_{ij}$ for the linear combination.
\begin{equation}
    \textbf{w}_{ij}=\frac{\textbf{I}^T((\textbf{I}\textbf{h}_{ij} - \textbf{H}_{ij})(\textbf{I}\textbf{h}_{ij} - \textbf{H}_{ij})^T)^{-1}}{\textbf{I}^T((\textbf{I}\textbf{h}_{ij} - \textbf{H}_{ij})(\textbf{I}\textbf{h}_{ij} - \textbf{H}_{ij})^T)^{-1}\textbf{I}}
\end{equation}

Afterward, the low-dimensional data can be reconstructed utilizing the weight $\textbf{w}_{ij}$, and the re-embedded data points should maintain the same relationship with the $c$-nearest neighbors that possess the same label. Then, the reconstruction error function is as follows, where $\hat{h}_{ij}$, $\hat{h}_{ijm}$ is the re-embedded representation for $h_{ij}$, $h_{ijm}$ in the new space.

\begin{equation}
   \underset{\hat{h}}{ \arg\min} \sum_{i=1}^{N} \sum_{j=1}^{k} \left\| \hat{h}_{ij}-\sum_{m=1}^{c} w_{ijm} \hat{h}_{ijm} \right\| ^{2}
\label{formula:reconstruction-error}
\end{equation}

Similar to the above process, 

\begin{equation}
\begin{aligned}
   f_{re} &= \left\| \hat{h}_{ij} - \sum_{m=1}^{c} w_{ijm} \hat{h}_{ijm} \right\| ^{2} \\
   &= \left\| \hat{\textbf{h}}_{ij} - \textbf{w}_{ij} \hat{\textbf{H}}_{ij} \right\| ^{2} \\
\end{aligned}
\label{formula:reconstruction}
\end{equation}

Let $\hat{\textbf{H}}=(\hat{\textbf{h}}_1, \hat{\textbf{h}}_2, ..., \hat{\textbf{h}}_{N \times k}) \in \mathcal{R}^{d^{\prime}\times (N \times k)}$, where $\hat{\textbf{h}}_i \in \mathcal{R}^{d^{\prime}\times 1}$ and $d^{\prime}$ is the dimension of re-embedded space. $\textbf{W}=(\textbf{w}_1, \textbf{w}_2, ..., \textbf{w}_{N \times k})^{T} \in \mathcal{R}^{d^{\prime}\times (N \times k)}$ where $\textbf{w}_i \in \mathcal{R}^{d^{\prime}\times 1}$. Let $ \textbf{M} = (\textbf{I}-\textbf{W})^{T}(\textbf{I}-\textbf{W})$, where $\textbf{I}$ is the identity matrix for brevity and Formula \ref{formula:reconstruction-error} can be rewritten as

\begin{equation}
    \underset{\hat{\textbf{H}}}{\min} \ \text{tr}(\hat{\textbf{H}}\textbf{M}\hat{\textbf{H}}^T)  \ \ \ \text{s.t.} \ \hat{\textbf{H}}\hat{\textbf{H}}^T = \textbf{I}
\label{formula:final}
\end{equation}

Formula \ref{formula:final} can be solved by eigenvalue decomposition, and the matrix of eigenvectors corresponding to the smallest $d^{\prime}$ eigenvalues is $\hat{\textbf{H}}^T$, which comprises the re-embedded representation.

\paragraph{Step 3.} During the inference for test instances, the representation of the $\mathtt{[MASK]}$ token $h_t$ is transformed into the new verbalizer space with the same dimension above. With Step 2, the weights for the linear combination are constructed with $c$-nearest neighbors of $h_{tm}, m \in 1,2,...,c$ from the training instances in the original space.

\begin{equation}
    \min  \left\| h_t - \sum_{m=1}^{c} w_{tm} h_{tm} \right\| ^{2}
\end{equation}

Then, the weights are employed to combine the re-embedded representation of the corresponding neighbors in Step 2 to transform $h_t$ into $\hat{h}_t$.

\begin{equation}
    \hat{h}_t = \sum_{m=1}^{c} w_{tm} \hat{h}_{tm}
\end{equation}

\paragraph{Step 4.} The procedure described above results in the representation of $\mathtt{[MASK]}$ tokens in the re-embedded space for both the training instances and test instances. The classification of a given test instance $\hat{h}_{t}$ is determined through its $e$-nearest neighbors $\hat{h}_{l}^{t}, l \in 1,2,..., e$ with the cosine distance from the training instances distributed in the re-embedded space.

\begin{equation}
    d(\hat{h}_{t}, \hat{h}_{l}^{t}) = \frac{\hat{h}_{t} \cdot \hat{h}_{l}^{t}}{\left\| \hat{h}_{t} \right\| \cdot \left\| \hat{h}_{l}^{t}\right\|}
\label{formula:dis}
\end{equation}

\begin{equation}
    P(y_{\hat{h}_{t}}=n | X=\hat{h}_{t}) = \frac{1}{e} \sum_{l=1}^{e} Ind(y_{\hat{h}_{l}^{t}}=n)
\end{equation}

\noindent where $Ind(\dots)$ represents an indicator function.

\subsection{Parameter Updating}
\label{parameter-updating}
Up until this point, no parameter has been tuned throughout the re-embedding process described above, thus avoiding the consumption of computational resources and the storage for the tuned models. The PTMs are regarded as a ``knowledge base'' \cite{petroni2019language} to some extent. 

However, it is possible that the PTMs do not fully grasp the tasks-specific information solely through pre-training. As a result, given the representation of the $\mathtt{[MASK]}$ tokens, we update the PTMs using contrastive learning by creating positive samples with instances from the same class and negative samples with instances from different classes, similar to the method in \cite{cui2022prototypical} as a plug-in module, yet without any additional encoder prior to the aforementioned re-embedding procedure, in order to provide the PTMs with task-specific knowledge as shown in Figure \ref{figure:model}.
Since a training instance can be represented by the embedding of the $\mathtt{[MASK]}$, denote $h_{ij},\ i\in1,2,...,N;j\in1,2,...,k$ as the set of training instances. 

A positive instance pair is formed by an instance $h_{ij}$ and another instance from the same class. A negative instance pair comprises two instances from different classes. The InfoNCE loss \cite{oord2018representation} is adopted as the contrastive learning loss and the loss function is as follows.

\begin{equation}
\mathcal{L}=-\frac{1}{kN}\sum_{i=1}^{N}\sum_{j=1}^{k}\log{\frac{\exp d({h}_{ij}, {h}_{in})}{\sum_{p \neq i}\exp d({h}_{ij}, {h}_{pq})}}
\end{equation}
where $d(\dots)$ is the distance metric in Formula \ref{formula:dis}.
\section{Experiments}

\begin{table*}[htpb]
\renewcommand{\arraystretch}{1.05}
\centering
\resizebox{0.9\textwidth}{!}{
\begin{tabular}{lccccc}
\hline \hline
            & \textbf{SST-2} (acc) & \textbf{MRPC} (F1)  & \textbf{QQP} (F1)  & \textbf{MNLI} (acc)   & \textbf{QNLI} (acc)         \\ \hline \hline
ManualVerb\dag & 54.6 (N/A) & 63.5 (N/A) & 63.2 (N/A)  & 32.3 (N/A) & 49.5 (N/A) \\
LLE-INC\dag    & 86.6 (0.8) & 67.6 (2.7) & 66.1 (6.3)  & 58.1 (3.5) & 62.4 (1.7) \\ \hline
Fine-tuning & 81.4 (3.8) & 76.6 (2.5) & 60.7 (4.3)  & 45.8 (6.4) & 60.2 (6.5) \\
ManualVerb  & 92.7 (0.9) & 74.5 (5.3) & 65.5 (5.3)  & 68.3 (2.3) & 64.5 (4.2) \\
GenVerb     & 92.3 (1.0) & \textbf{76.2} (2.3) & 67.0 (3.0)  & 68.3 (2.5) & 68.3 (7.4) \\
ProtoVerb   & 87.8 (3.1) & 66.9 (1.5) & 66.9 (7.0)  & 62.1 (3.2) & 59.9 (4.8) \\
LLE-INC     & \textbf{92.9} (2.4) & \textbf{76.2} (3.1) & \textbf{68.1} (8.1)  & \textbf{69.2} (1.9) & \textbf{70.2} (4.1) \\ 
 \hline \hline
            & \textbf{RTE} (acc)  & \textbf{CHIP-CTC} (acc) & \textbf{cMedTC} (acc) & \textbf{Kuake-QIC} (acc) & \textbf{Tnews} (acc) \\ \hline
ManualVerb\dag  & 47.7 (N/A) & 9.4 (N/A)    & 10.4 (N/A)    & 2.3 (N/A)    & 6.1 (N/A)    \\
LLE-INC\dag     & 68.2 (4.1) & 44.5 (2.2) & 47.7 (2.4)  & 31.2 (2.9) & 41.3 (2.6) \\ \hline 
Fine-tuning & 54.4 (3.9) & 29.0 (7.2) & 25.2 (5.7)  & 13.1 (8.7) & 13.5 (2.4) \\
ManualVerb  & 69.1 (3.6) & 57.8 (1.8) & 62.9 (3.2)  & 52.8 (3.4) & 43.9 (2.2) \\
GenVerb     & 73.9 (2.2) & 32.3 (1.9) & 19.3 (14.2) & 44.9 (2.9) & 7.4 (4.6)  \\
ProtoVerb   & 70.1 (3.7) & 51.7 (2.4) & 53.2 (7.7)  & 48.0 (0.9) & 40.1 (1.5) \\
LLE-INC     & \textbf{74.5} (1.9) & \textbf{61.0} (3.1) & \textbf{63.0} (1.2)  & \textbf{54.8} (1.3) & \textbf{44.6} (1.1) \\ \hline
\hline
\end{tabular}
}
\caption{Results with various prompt verbalizers. We report the mean (standard deviation) performance of accuracy/F1 over 5 random seeds. \dag: tuning-free. N/A: not applicable. Bold: best results.}
\label{main}
\end{table*}

\begin{table*}[t]
\renewcommand{\arraystretch}{1.05}
\centering

\resizebox{0.85\textwidth}{!}{
\begin{tabular}{lccccccccc}
\hline
\hline
& \#p & \#t-p  & \textbf{SST-2} & \textbf{MRPC}   & \textbf{QQP} & \textbf{MNLI} &  \textbf{QNLI}  & \textbf{RTE}  \\ \hline
RoBERTa-LLE-INC  & 355M & 355M   & 92.9  & 76.2  & 68.1  & 69.2  & 70.2  & 74.5 \\ \hline

LLaMA-LLE-INC\dag  & 7B & 0  &  93.2   &  81.4   & 77.3 & 66.5 & 73.2 & 74.5   \\
LLaMA-w/o re-embed\dag   & 7B & 0 & 88.0  &   79.3 &  70.3    & 57.0      &  71.2     & 72.2      \\    \hline

LLaMA-LLE-INC\dag & 13B & 0 &\textbf{93.8}&82.9&79.2 &70.1 & 76.2        & 79.1        \\

LLaMA-w/o re-embed\dag   & 13B & 0 & 88.8  &   80.6   &  75.6    & 64.9   &  73.9  & 73.0 \\    \hline

LLaMA-LLE-INC\dag & 65B & 0   &92.8&\textbf{83.0}&\textbf{81.7} &\textbf{73.5} & \textbf{84.3}        & \textbf{83.4}        \\

LLaMA-w/o re-embed\dag  & 65B & 0   & 86.8  &   81.2   & 77.4    & 66.2   &   79.1 & 73.6 \\
 \hline \hline
\end{tabular}
}

\caption{Experimental results for LLaMA-7B, LLaMA-13B and LLaMA-65B with LLE-INC. \#p: the number of parameters. \#t-p: the number of tuned parameters. \dag: tuning-free.}
\label{table:ablation-llama}
\end{table*}

\subsection{Datasets}
We conduct experiments to demonstrate the effectiveness of our approach with 10 classification datasets (including 4 multi-class datasets) in both English and Chinese from GLUE \cite{wang2018glue}, CLUE \cite{xu2020clue} and CBLUE \cite{zhang2022cblue} benchmarks. The datasets include 6 datasets from GLUE: SST-2 \cite{socher2013recursive}, MRPC \cite{dolan2005automatically}, QQP \footnote{\url{https://www.quora.com/q/quoradata/}}, QNLI \cite{rajpurkar2016squad} and RTE \cite{dagan2006pascal}, 3 datasets from CBLUE: CHIP-CTC \cite{zhang2022cblue}, cMedTC \cite{zhang2020conceptualized} and KUAKE-QIC \cite{zhang2022cblue} and 1 dataset from CLUE: Tnews \cite{xu2020clue}. In particular, the datasets from CBLUE and CLUE are multi-class datasets. The statistics of the datasets are presented in Table \ref{table:data_stat} in Appendix \ref{appendix:stats}. For the GLUE benchmark, we follow Gao et al. \cite{gao-etal-2021-making} to use the original development sets as test sets and randomly select 16 instances for each class from the training set using 5 random seeds in few-shot scenarios and test with the full-size test set. 

\subsection{Baselines}
We report our experimental results compared with the baselines with various verbalizer design methods, including manual verbalizer, automatic verbalizer and prototypical verbalizer. For a fair comparison, we fix the prompt template across various verbalizer methods and just investigate the performance with different verbalizer designs. We also experiment with the standard \textbf{Fine-tuning}. Manual Verbalizers (\textbf{ManualVerb}) are selected manually by domain experts empirically. Generated Token Verbalizers (\textbf{GenVerb}) are automatically searched from the vocabulary list of the PTMs. Here, we adopt the approach in LM-BFF \cite{gao-etal-2021-making}, which uses the conditional likelihood and re-ranking strategy to find the optimal token for the verbalizers.
Prototypical Verbalizers (\textbf{ProtoVerb}) \cite{cui2022prototypical} are prototype embeddings directly learned for each class from the representation of training samples. During the inference process, the PTM makes the prediction by measuring the similarity between the query and every prototype embedding.

We implement the baselines with the Huggingface Transformers \cite{wolf2020transformers} package based on PyTorch \cite{paszke2019pytorch} framework and OpenPrompt \cite{ding2022openprompt} toolkit. We adopt the RoBERTa-Large model \cite{liu2019roberta, cui2021pre} as our backbone PTM for both the English and Chinese tasks. Implementation details including prompt settings and experiment settings are in Appendix \ref{appendix:implementation}.

\subsection{Parameter Updating}
We introduced LLE-INC, a method that does not require additional parameters or parameter updating of the PTMs. In contrast, the baselines have different requirements. Vanilla Fine-tuning demands the training of a linear classifier, and GenVerb entails the exploration of optimal verbalizers and updating the PTMs with the verbalizers. ProtoVerb, although able to operate with frozen PTMs, necessitates the training of additional prototypical embeddings. Only ManualVerb is capable of directly operating on tasks following the pre-training process without tuning for downstream tasks. Consequently, we compare LLE-INC with the baselines under two conditions: one \textit{with} parameter tuning and one \textit{without} parameter tuning (including PTMs and additional parameters).

\subsection{Results}
We report the mean accuracy/F1 score following the baselines for each task across 5 sampled few-shot datasets using various random seeds, along with the standard deviation. The results for the experiments with and without parameter tuning are shown in Table \ref{main}.

\paragraph{Without Parameter Updating} LLE-INC re-embeds the $\mathtt{[MASK]}$ representation without requiring the addition of any new parameters or parameter updates as introduced. In this situation, we only employ the frozen parameters in the PTM. Without any tuning parameters, only the ManualVerb in the baselines can still make predictions. The ManualVerb and LLE-INC which tune no parameter are denoted as ManualVerb\dag \ and LLE-INC\dag, respectively. Table \ref{main} shows that ManualVerb\dag \ performs poorly without tuning, which means the model cannot understand the tasks and manually designed verbalizers explicitly. The performance of ManualVerb\dag \ on the test set is constant across different random seeds since the PTM is frozen. LLE-INC\dag, on the other hand, predicts the labels far more accurately than ManualVerb\dag \ and is on par with or even sometimes outperforms GenVerb and ProtoVerb (both demand parameter updating) because the space reconstruction of LLE-INC is based on the relationship with the intra-class neighbors. The aforementioned finding suggests that the output embeddings from the frozen PTM contain a significant amount of implicit information and manifold-based re-embedding is a viable approach to leverage the information.

\paragraph{With Parameter Updating}
Since the PTMs may not completely understand the task information, we can first train the models using contrastive learning. Then, we utilize the LLE-INC to re-embed the output representation of the updated PTMs as illustrated in Figure \ref{figure:model}. Table \ref{main} demonstrates that LLE-INC outperforms the baselines of Fine-tuning, GenVerb, ProtoVerb and even ManualVerb consistently and the performance improvements are statistically significant with the p-value of paired t-test less than 0.05 in the majority of all cases. Our approach can bring up to 3.2\% improvement on the CHIP-CTC dataset and 1.1\% improvement on average across the 10 datasets. The experimental results of LLE-INC\dag \ reveal that the potential of PTMs can be reached by the manifold-based embedding space re-construction and the performance of LLE-INC further demonstrates that contrastive learning-based tuning for the parameters in the PTMs integrates better with the space re-embedding. As it comes to harder multi-class tasks, LLE-INC shows superiority to the baseline verbalizers without human effort for the re-embedding process can yield better representation for the instance. Meanwhile, we also notice that there can be duplicate verbalizers generated by GenVerb while dealing with multi-class datasets which negatively affects its performance. 

\subsection{Large Language Model with LLE-INC}
The emergence of large language models (LLMs) has significantly transformed the field of natural language processing (NLP), leading to superior performance compared to earlier-generation paradigms. However, the fine-tuning of LLMs can pose a significant challenge due to the scale of their parameters. To address this challenge, we apply the LLE-INC to the tuning-free outputs of the LLaMA models \cite{touvron2023llama} (LLaMA-7B, LLaMA-13B, and LLaMA-65B) to explore the performance of LLE-INC. As LLaMA is primarily trained on English corpora, we only consider English language tasks. Our results, presented in Table \ref{table:ablation-llama}, show that LLaMA-LLE-INC\dag, without any tuning parameters, can achieve better performance than RoBERTa-large with the PTM updating. This indicates that LLE-INC can be applied as a tuning-free method for classification tasks using large language models.

\section{Analysis}
\paragraph{Re-embedding Strategy}
LLE-INC re-embeds the verbalizer embedding space under the constraint of intra-class neighbors and makes predictions for the test instances with the k-nearest neighbors in the re-embedded space. Therefore, we also experiment with other re-embedding strategies: (1) LLE, the original locally linear embedding which only relies on the k-nearest-neighbor spatial relationship in the original space, and (2) w/o re-embedding, which makes predictions on the embeddings in the original space only with a kNN classifier. Experiments here do not include the contrastive learning module for a fair comparison. Table \ref{table:ablation} shows that the performance of w/o re-embedding\dag is inferior to that of LLE\dag \ and LLE-INC\dag, indicating that the performance is limited by the original space. Furthermore, LLE\dag \ performs much better than w/o re-embedding\dag \ (up to 12.7\% improvement) and LLE-INC\dag \ outperforms LLE\dag \ (up to 7.9\% improvement) and demonstrates that the intra-class neighbor relationship is superior to spatial $k$-nearest neighbors.

\paragraph{Ablation Study}
An ablation study was conducted to evaluate the effectiveness of contrastive learning and space reconstruction. The results in Table \ref{table:cl-ablation} demonstrate that the performance of LLE-INC in combination with contrastive learning surpasses that of either approach individually. This indicates that both parameter updating and embedding space reconstruction contribute to improved model performance.
\begin{table}[tbp]
\renewcommand{\arraystretch}{1.05}
\centering

\resizebox{0.99\columnwidth}{!}{
\begin{tabular}{lccc}
\hline
\hline

& \textbf{SST-2} & \textbf{MRPC}   & \textbf{QQP}      \\ \hline
LLE-INC\dag       & \textbf{86.6}      &   \textbf{67.6}       &  \textbf{66.1}                       \\
LLE\dag        & 79.2      &   61.4       &  62.1                 \\
w/o re-embed\dag       & 66.5      &   54.3       &  60.9                     \\ 

\hline \hline

& \textbf{MNLI}&  \textbf{QNLI} & \textbf{RTE}     \\ \hline
LLE-INC\dag   & \textbf{58.1}    &  \textbf{62.4} &   \textbf{68.2}                  \\
LLE\dag    & 51.9          &  58.0 &   60.3           \\
w/o re-embed\dag  & 42.4    &  54.7     &   61.1                        \\ 

\hline \hline

& \textbf{CHIP-CTC} & \textbf{Kuake-QIC} &\textbf{Tnews} \\ \hline
LLE-INC\dag     & \textbf{44.5}    & \textbf{31.2} & \textbf{41.3}                       \\
LLE\dag    & 39.2         & 29.6 & 39.5                  \\
w/o re-embed\dag   & 33.8    & 28.1 & 40.3               \\ \hline \hline
\end{tabular}
}

\caption{Experimental results for different re-embedding strategies. \dag: tuning-free.}
\label{table:ablation}
\end{table}

\begin{table}[h]
\renewcommand{\arraystretch}{1.05}
\centering
\resizebox{0.99\columnwidth}{!}{
\begin{tabular}{lccc}
\hline
\hline
& \textbf{SST-2} & \textbf{MRPC}   & \textbf{QQP}        \\ \hline
\small{LLE-INC with CL}  & \textbf{92.9}   &   \textbf{76.2}       &  \textbf{68.1}                  \\
\small{LLE-INC}             & 86.6    &           67.6        &          66.1                     \\
\small{CL}             & 85.1    &           64.2        &          66.5                     \\ \hline \hline

& \textbf{MNLI} &  \textbf{QNLI} & \textbf{RTE}   \\ \hline
\small{LLE-INC with CL}     & \textbf{69.2}   &  \textbf{70.2}    & \textbf{74.5}  \\
\small{LLE-INC}       & 58.1            &  62.4            &   68.2    \\
\small{CL}         & 57.0            &  60.4         &   67.8    
\\ \hline \hline

 & \textbf{CHIP-CTC} & \textbf{Kuake-QIC} &\textbf{Tnews}  \\ \hline
\small{LLE-INC with CL}        & \textbf{61.0}   & \textbf{54.8} & \textbf{44.6}           \\
\small{LLE-INC}                       & 44.5             & 31.2      &  41.3                 \\
\small{CL}                     & 49.7              & 40.1      &  38.2                    
\\ \hline \hline
\end{tabular}
}

\caption{Ablation study for the LLE-INC and contrastive learning. CL: contrastive learning.}
\label{table:cl-ablation}
\end{table}

\section{Conclusion}
Prompt-based tuning has been proven effective in few-shot scenarios and recently embedded verbalizers have been explored as an alternative to the labor-intensive process of existing verbalizers. Recent studies are dependent on the tuning of the PTM or extra trainable embeddings and the manifold in the high-dimensional representation space has the potential to mislead Euclidean distance measurements. In this study, we propose to re-embed the verbalizer representation space through locally linear embedding with an intra-class neighborhood constraint. Experimental results demonstrate that LLE-INC works rather well without any parameter tuning and can further enhance the performance of prompt-based tuning in conjunction with model tuning. The effectiveness of manifold learning on the LLaMA model also opens up new possibilities for the tuning-free application of LLMs and further inspires computational resource-friendly research on the LLMs.

\clearpage
\newpage
\section{Acknowledgements}
We thank the anonymous reviewers for their insightful and constructive comments and gratefully acknowledge the support of the National Key R\&D Program of China (2021ZD0113302), the National Natural Science Foundation of China Youth Fund (62206079), and the Heilongjiang Provincial Natural Science Foundation of China (YQ2022F006). We also appreciate the support from Du Xiaoman (Beijing) Science Technology Co., Ltd. on our research.
\bibliography{bib}
\clearpage
\newpage
\appendix
\section{Notation Table for Variables and Parameters} \label{appendix:notation}
The notation for the variables and parameters in this paper can be found in Table \ref{table:notation}.
\begin{table*}[b]
\centering

\begin{tabular}{lll}
\hline \hline
\textbf{Notation} & \textbf{Description} \\
\hline
$h_{\mathtt{[MASK]}}$ & the hidden state embedding for the [MASK] token   \\
$h_{\text{instance}}$ & the representation for an input instance  \\
$i\in1,2,...,N$ & N-way in few-shot settings, the index for the class  \\
$j\in1,2,...,k$ & k-shot in few-shot settings, the index for the instance in a class  \\
$h_{ij}$ & the representation for a training instance \\
$m\in1,2,...,c$ & the index for the c-nearest neighbors for a training instance $h_{ij}$ \\
$h_{ijm}$ & the representation for the c-nearest neighbors for a training instance $h_{ij}$ \\
& in the same class \\
$w_{ijm}$ & the weight for $h_{ijm}$ \\
$d$ & the dimension of embeddings in the original space \\
$d^{\prime}$ & the dimension of embeddings in the re-embedded space \\
$\textbf{h}_{ij}$ & vector format of $h_{ij}$, $\textbf{h}_{ij} \in \mathcal{R}^{1\times d}$ \\
$\textbf{w}_{ij}$ & vector format of $w_{ij}$, $\textbf{w}_{ij} \in \mathcal{R}^{1\times c}$ \\
$\textbf{H}_{ij}$ & matrix format of $h_{ijm}$, $\textbf{H}_{ij} \in \mathcal{R}^{c\times d}$ \\
$\hat{h}_{ij}$ & the re-embedded representation for $h_{ij}$ \\
$\hat{h}_{ijm}$ & the re-embedded representation for $h_{ijm}$ \\
$\hat{\textbf{H}}_{ij}$ & the re-embedded representation for $\textbf{H}_{ij}$ \\
$\hat{\textbf{H}}$ & the matrix comprising the re-embedded representation of all training instances \\
 & $\hat{\textbf{H}}=(\hat{\textbf{h}}_1, \hat{\textbf{h}}_2, ..., \hat{\textbf{h}}_{N \times k}) \in \mathcal{R}^{d^{\prime}\times (N \times k)}$\\
$h_{t}$ & the representation for a test instance \\
$h_{tm}$ & the representation for the c-nearest neighbors for a test instance $h_{t}$ \\
& in the original space \\
$w_{tm}$ & the weight for $h_{tm}$ \\ 
$\hat{h}_{t}$ & the re-embedded representation for $h_{t}$ \\
$\hat{h}_{tm}$ & the re-embedded representation for $h_{tm}$ \\
$l\in1,2,...,e$ & the index for the e-nearest neighbors for a test instance $\hat{h}_{t}$ \\
$\hat{h}_{l}^{t}$ & the representation for the e-nearest neighbors for a $\hat{h}_{t}$ in the re-embedded space \\ 
\hline \hline
\end{tabular}
\caption{
Description of the variables and parameters involved in the formulas.}
\label{table:notation}
\end{table*}

\section{Statistics of Datasets} \label{appendix:stats}
The statistics of the datasets are in Table \ref{table:data_stat}.
\begin{table*}[htbp]
\centering

\begin{tabular}{lccccc}
\hline
\hline
\textbf{Dataset}  & \textbf{Benchmark}  & \textbf{Language} & \textbf{Task} & \textbf{\#Class}   & \textbf{\#Test} \\
\hline
SST-2 & GLUE & English & Sentiment Classification & 2 & 872 \\
MRPC & GLUE & English & Paraphrase & 2 & 408\\
QQP & GLUE & English & Paraphrase & 2 & 40431 \\
MNLI & GLUE & English & Natural Language Inference & 3 & 9815 \\
QNLI & GLUE & English & Natural Language Inference & 2 & 5463 \\
RTE  & GLUE & English & Natural Language Inference & 2 & 277 \\
CHIP-CTC & CBLUE & Chinese & Clinical Trial Classification & 22 &  7,682 \\
cMedTC  & CBLUE   & Chinese & Clinical Trial Classification & 16 &  1,800 \\
KUAKE-QIC & CBLUE  & Chinese & Query Intention Classification & 11 & 1,955 \\
Tnews  & CLUE   & Chinese & Topic Classification & 15 & 10,000 \\
\hline \hline
\end{tabular}
\caption{Statistics of the datasets.}
\label{table:data_stat}
\end{table*}

\section{Implementation} \label{appendix:implementation}
\paragraph{Prompt Settings}
For a fair comparison, we use the same prompt settings including the prompt templates and verbalizers in Gao et al. \cite{gao-etal-2021-making} for the ManualVerb and GenVerb with the GLUE benchmark datasets. For the datasets in Chinese, we adopt the prompt template ``It is about $\mathtt{[MASK]}$.'' (in Chinese) and the optimal single character in the task labels as prompt verbalizers through experiments. For the LLaMA model, we append ``It is'' to the inputs for the tasks with single sentences and adopt ``Sentence 1: [sentence 1]. Sentence 2: + [sentence 2]. Sentence 1 and Sentence 2 are'' for the tasks with double sentences. Then, we get the hidden representation of the first generated token as the representation for a specific instance. 
\paragraph{Experiment Settings} The AdamW \cite{loshchilov2018decoupled} is adopted for model optimization. We set the hyper-parameters as the baseline methods reported, and for those without reported hyper-parameters, we set the learning rate to 3e-5, the batch size to 16, and the epoch to 5. For the contrastive learning, we set the learning rate to 3e-5, the temperature to 0.05, and the batch size to twice the number of classes in the datasets during the contrastive learning. For the LLE and LLE-INC, we re-embed the space into the dimension from 20 to 400 with all possible intra-class neighbors for various tasks. 
\paragraph{Computational resources} LLE-INC is run on CPUs and without the requirement of GPU. The experiments involving the PTMs are run on single NVIDIA Tesla V100 or A100 GPUs. 

\section{Limitations}
Our study re-embeds the prompt verbalizer space by leveraging the local neighborhood feature within the same class and LLE-INC can performance fairly well with tuning \textit{no} parameter. The dimension of the re-embedded space is a hyper-parameter, similar to the batch size and learning rate, and the optimal dimensions fall into the range of 20 to 400 for the RoBERTa-large and the LLaMA models. It can be worthwhile to exploit the determination of the optimal dimension in future studies.

\end{document}